\pdfoutput=1

\documentclass[11pt]{article}

\usepackage[]{ACL2023}

\usepackage{times}
\usepackage{latexsym}

\usepackage[T1]{fontenc}

\usepackage[utf8]{inputenc}

\usepackage{microtype}

\usepackage{inconsolata}
\usepackage{graphicx}
\usepackage{tabularray}
\usepackage{multirow}
\usepackage{makecell}
\usepackage{amsmath}
\usepackage{enumitem}
\usepackage{booktabs}
\usepackage{xcolor}
\definecolor{mint}{RGB}{50,255,255}
\usepackage{float}
\usepackage{csquotes}

\makeatletter
\def\hlinewd#1{%
\noalign{\ifnum0=`}\fi\hrule \@height #1 %
\futurelet\reserved@a\@xhline}
\makeatother
\title{Instruct-SCTG: Guiding Sequential Controlled Text Generation through Instructions}

\author{
 \textbf{Yinhong Liu}$^\spadesuit$  \quad \quad
 \textbf{Yixuan Su}$^\spadesuit$  \quad \quad 
 \textbf{Ehsan Shareghi}$^{\heartsuit \spadesuit}$ \quad \quad 
 \textbf{Nigel Collier}$^\spadesuit$  \quad
 \\
 $^\spadesuit$Language Technology Lab, University of Cambridge\\
 $^\heartsuit$Department of Data Science and AI, Monash University\\
 {\tt \{yl535,ys484,nhc30\}@cam.ac.uk}\\
 {\tt ehsan.shareghi@monash.edu}
}

\begin{document}
\maketitle
\begin{abstract}

Instruction-tuned large language models have shown remarkable performance in aligning generated text with user intentions across various tasks. However, maintaining human-like discourse structure in the generated text remains a challenging research question. 
In this paper, we propose Instruct-SCTG, a flexible and effective sequential framework that harnesses instruction-tuned language models to generate structurally coherent text in both fine-tuned and zero-shot setups. 
Our framework generates articles in a section-by-section manner, aligned with the desired human structure using natural language instructions. 
Furthermore, we introduce a new automatic metric that measures discourse divergence in a fuzzy manner. 
Extensive experiments on three datasets from representative domains of news and recipes demonstrate the state-of-the-art performance of our framework in imposing discourse structure during text generation, as verified by both automatic and human evaluation.
Our code will be available on Github.

\end{abstract}


\section{Introduction}

\begin{figure*} [th!]
\begin{center}
\includegraphics[width=1.0\linewidth]{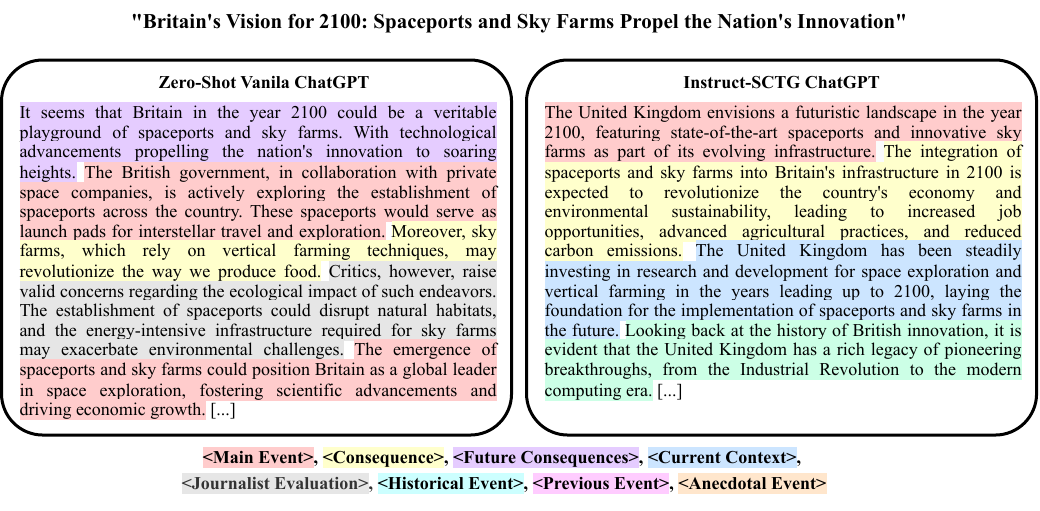}%
\end{center}
\vspace{-3mm}
\caption{Comparing examples with discourse role labels: left (zero-shot ChatGPT) vs. right (Instruct-SCTG framework utilizing zero-shot ChatGPT as backbone generator). The right article exhibits improved content flow and enhanced discourse structure.}
\label{fig:example_compare}
\vspace{-3mm}
\end{figure*}

The recent progress in Language Models (LMs) have attracted widespread attention from both academia and industry. 
These models, powered by massive corpora and advanced hardware, have demonstrated improving performance across various NLP benchmarks, ranging from generative tasks, such as Machine Translation or Data-to-Text generation, to understanding tasks, e.g. GLUE~\citep{wang-etal-2018-glue}.
In particular, Large Language Models (LLMs) designed for instruction-following, such as ChatGPT\footnote{https://openai.com/blog/chatgpt} and Flan-T5~\citep{chung2022scaling}, exhibit impressive capabilities in comprehending instructions expressed in natural language and precisely aligning the model outputs with human intentions.

Generating high-quality text is essential for various Natural Language Generation (NLG) tasks. However, certain tasks, such as news report generation, require more than just textual fluency. 
Effectively organizing the underlying discourse structure of an article can help readers quickly grasp key information, enhancing engagement and readability. 
For example, an experienced journalist can coherently structure the core event, background, consequence, critics' evaluations and other elements of a news report. 
As shown in Fig.~\ref{fig:example_compare}, a well-structured report can efficiently deliver event information, capture readers' attention and even convey opinions.
The task of text generation with specific discourse structure constraints has long been a research focus in the field covering various domains, including stories, news, recipes and question answering. 
We address this challenge as the task of Sequential Controlled Text Generation (SCTG), previously formulated by  \citet{spangher-etal-2022-sequentially}. 
In SCTG, the goal is to generate coherent text following an input prompt and a sequence of control code.

In this paper, we propose Instruct-SCTG, a simple yet effective framework that harnesses instruction-following LMs to generate structurally coherent text. 
Specifically, our framework breaks down the generation task into a sequence of sub-tasks and guides the Supervised Fine-tuned (SFT) LMs sequentially to produce content section by section through natural language instructions.
This approach effectively aligns the resulting articles with the given discourse structures, enhancing the overall coherence and readability of the generated text.
We also investigate crucial factors to consider during the SFT stage, such as different levels of discourse information exposure.
Furthermore, to evaluate the adherence of generated articles to the input control codes, we introduce a novel automatic metric that measures discourse divergence in a fuzzy positional manner.

We conducted extensive experiments using three datasets from two representative domains, i.e. news and recipes. 
For news articles, we utilized the All-The-News dataset\footnote{kaggle.com/snapcrack/all-the-news.} from Kaggle and the News Discourse dataset \citep{choubey-etal-2020-discourse}. 
For recipe generation, the experiments were performed on the Recipe1M+ dataset \citep{marin2019learning}. 
We assess the textual fluency and structural coherence of the generated text with both automatic and human evaluations. 
The results demonstrate the effectiveness of our framework in controlling LMs to generate text adhering to the given discourse structures.

In summary, our contributions are three-folds: 
Firstly, we introduce a straightforward yet effective framework that leverages instruction-following LMs to generate structurally coherent texts in the task of SCTG, achieving state-of-the-art (SOTA) performance on three datasets from two representative domains.
Secondly, we introduce a novel automatic metric that can effectively measure the fuzzy adherence of discourse structure.
Lastly, our work is the first one that explore the design of instructions to exert control over the underlying discourse structure during text generation.

\section{Background and Related Works}
\subsection{Instruction Fine-tuned Language Model}

Instruction-following LMs are language models specially optimized to comprehend and execute natural language instructions.
These models leverage large-scale Pre-trained Language Models (PLM) like GPT-3 and incorporate an additional supervised aligning fine-tuning process. 
Their recent emergence has significantly advanced the understanding of human intentions and the generation process conditioning on those intentions. 

For instance, InstructGPT~\citep{ouyang2022training} fine-tunes GPT-3 \citep{brown2020language}
to achieve human desired model behavior through reinforcement learning from human feedback (RLHF, \citet{christiano2017deep, stiennon2020learning}).
Similarly, Flan-T5 \citep{chung2022scaling} fine-tunes the T5 language model \citep{raffel2020exploring} using a diverse range of instruction templates from a collection of data sources.
Another example is Alpaca, proposed by \citet{alpaca}, which is an instruction fine-tuned Language model based on LLaMA~\citep{touvron2023llama}, using an instruction dataset generated in the style of self-instruct \citep{selfinstruct}.
These instruction-following LLMs showcase the progress in leveraging instructions to guide language generation, facilitating a more interactive and controllable generation process.

\subsection{Discourse Structure}


Discourse structure investigates the organization of language into larger units like paragraphs, sections, and complete articles. 
In this work, we focus on the communicative functions within entire articles served by those linguistic units.
Therefore, texts from different domains are characterized by different discourse schemas, as their linguistic units also play different functional roles.
The discourse roles of scientific papers or experimental abstracts~\citep{liddy1991discourse,mizuta2006zone} include background, methodology, experiments and findings.
In the domain of long-form question answering \citet{xu2022we}, the discourse function of each sentence can be answer, summary, example and so on. 
\citet{liu-etal-2022-plug} developed a discourse schema for recipes based on actions and controlled the generation process according to the predicted discourse sequences. 
The explicit functional discourse structure of news reports was addressed~\citep{van2013news, choubey-etal-2020-discourse} by defining roles based on their relations with the main event, such as consequence and journalist evaluation.

Multiple established frameworks also proposed different definition of discourse structure, which focus on how each linguistic unit relates to each other through discourse connectives, such as causal, temporal, etc.
For instance, Rhetorical Structure Theory, RST \citep{mann1988rhetorical}, seeks to identify rhetorical relations between text segments and form a hierarchical organization of discourse.
The Penn Discourse Treebank, PDTB \citep{prasad2008penn}, defines its schema based on low-level discourse connectives presented in the text.

\subsection{Sequential Controlled Text Generation}

Extensive research has been conducted on Controlled Text Generation (CTG) to enable the control of attributes such as lexical constraints, style and length in the output of PLM. 
One notable example is prefix-tuning, introduced by \citet{li2021prefix}, which only optimizes a short task-specific vector (prefix) while keeping the rest of the PLM frozen, thereby controlling the domain of generation. 
Another representative work is PPLM by \citet{dathathriplug}, which uses gradients from an attribute discriminant model to steer the text generation.

In this work, we focus specifically on the task of Sequential Controlled Text Generation (SCTG), recently formalized by \citet{spangher2023sequentially}.
In SCTG, a model is provided with an input prompt and a sequence of control codes, and the output is a text sequence comprising multiple sentences. 
Each control code specifies the desired content or style of the corresponding output sentence, enabling control over the inter-sentence structure of the generated text. 
The task of SCTG is different from the conventional CTG tasks, which focuses on controlling isolated local attributes at a time. However, SCTG tackles a more intricate challenge. The generation conditions not only on the discourse of the current sentence or paragraph but also on previous text and contextual discourse structure to maintain contextual coherence throughout the articles.

Previous works relevant to this task include \citet{liu-etal-2022-plug}, who proposed a plug-and-play guided decoding method that predicts content plans to control the generation process accordingly.
For coherent text generation that considers discourse, \citet{bosselut2018discourse} modeled discourse structure as cross-sentence ordering.
Furthermore, \citet{spangher2023sequentially} introduced a pipeline method that improves discourse through guided generation and an overall editing process.

\begin{figure*} [ht!]
\begin{center}
\hspace*{-2mm}
\includegraphics[width=0.8\linewidth]{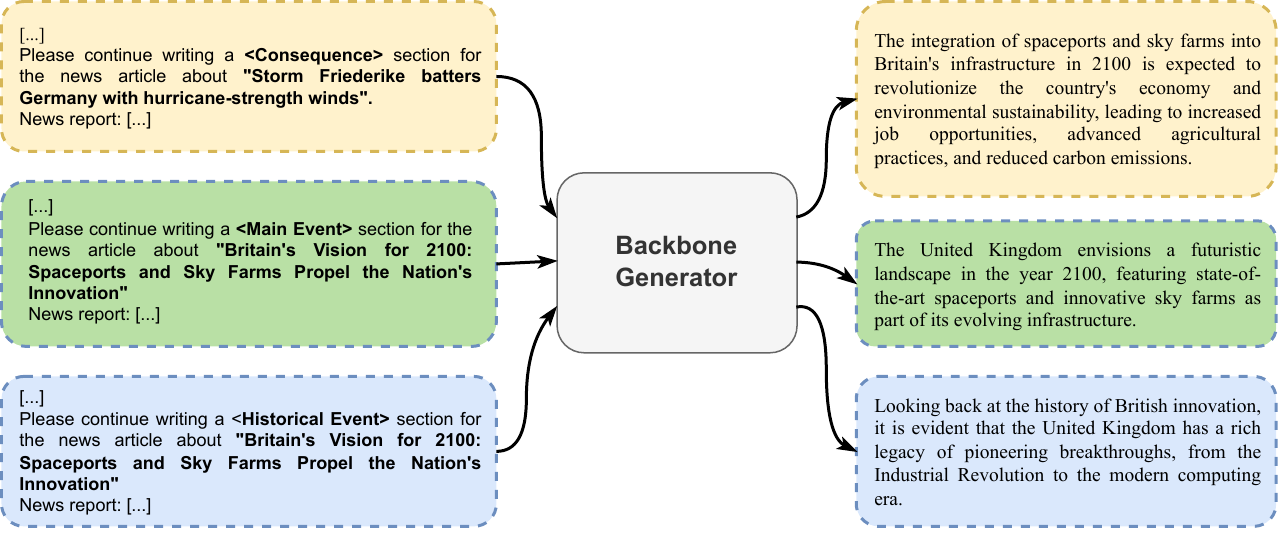}%
\end{center}
\caption{Overview of the instruction tuning of the backbone generator for the Instruct-SCTG.}
\label{fig:overview}
\vspace{-3mm}
\end{figure*}

\section{Methodology}

\subsection{Overview}
We propose a novel framework called Instruct-SCTG (Instruction Sequential Control Text Generation) to incorporate discourse structure into generated articles, by decomposing the generation process into a series of sub-tasks. 
Each sub-task is designed to generate a single specific text section, such as a main event section or journalist evaluation section, based on the given discourse sequence. 
In this section, we explain our framework in details and how we design the SFT instruction for our generator LM.
Additionally, we introduce an automatic metric that measures the adherence of the discourse structure.

\subsection{Instruct-SCTG Framework}
\label{sec:seq}
\textbf{Task Formulation.} 
The goal of SCTG is to generate a coherent article represented by a sequence of linguistic units, e.g. sentences, $\boldsymbol{x} = \{x_1, ..., x_{|\boldsymbol{x}|}\}$. 
Each unit $x_i$ is denoted as $x_i=\{x_{i,1}, ..., x_{i,|x_i|}\}$, where $x_{i,j}$ is the $j$-th token of $x_i$.
In the formulation of SCTG, we assume that the input information $\mathcal{I}$, such as news headlines or recipe title and ingredients, and discourse structure are provided.
The discourse structure is represented as a control code sequence $\boldsymbol{c}= \{c_1, ..., c_s\}$, where each code denotes the expected discourse role for its corresponding unit $x_s$.
Hence, the objective of generation is to model the conditional distribution of the document $\boldsymbol{x}$, expressed by Equation~\ref{equ:cond-prob}.
\begin{equation}\label{equ:cond-prob}
    P(\boldsymbol{x}|\boldsymbol{c}, \mathcal{I}) = \prod_{i=1}^s p(x_i| \boldsymbol{x}_{<i}, \boldsymbol{c}, \mathcal{I})
\end{equation}
\textbf{Our Framework.} 
We decompose the document-level conditional distribution into a series of unit-level sub-tasks. 
During each iteration, we instruct the backbone generator to continue writing for the current linguistic unit according to the specified control code.
We can use either a fine-tuned LM with task-specific instructions or a zero-shot large LM as the backbone generator. 
The control code $\boldsymbol{c}$, or discourse roles, are predefined categories based on a specific discourse schema designed for different domains and tasks.
In Section~\ref{sec:verbalization}, we explain the design of our SFT instructions.

\subsection{Task-specific Instruction Tuning}
\label{sec:verbalization}
To prepare the backbone generator for our sequential framework, we design task-specific instructions for fine-tuning LMs.
As shown in Figure~\ref{fig:overview}, our approach segments articles into sentences or paragraphs. We then create instruction--paragraph pairs as the Supervised Fine-tuning data.

In this section, we also explore the impact of different instruction designs on the resulting fine-tuned generator.
The instructions, as shown in the example in Table~\ref{tab:instruction_template}, consists of three main components: (i) discourse context, (ii) input information and (iii) textual context. 

In the discourse context, we specifically explore the influence of various facets of contextual discourse information on the generator's control performance. 
While exposure to extensive discourse context offers more information, it can potentially introduce additional noise to the current generation process. 
Previous research~\citep{spangher-etal-2022-sequentially} employed three levels of discourse dependency assumptions (local, past-aware and Full-sequence)  when setting up the discriminator in their post-processing controlling algorithm. 
In contrast, in this work, we include diverse levels of discourse context in our instructions.
This variation enables us to simulate the those dependency approximations, such that we can directly  condition the text generation process on them. 

\paragraph{Local discourse.} If we assume the generation of the current linguistic unit only depends on its corresponding discourse role, but not the contextual discourse structure, the conditional distribution Equation~\ref{equ:cond-prob} can be simplified as below.
\begin{align*}\label{equ:local}
   P(\boldsymbol{x}|\boldsymbol{c}, \mathcal{I}) \approx & \prod_{i=1}^{|\boldsymbol{x}|} \prod_{j=1}^{|x_i|} p(x_{i,j} | x_{i,<j}, \boldsymbol{x}_{<i}, , c_{i}, \mathcal{I})
\end{align*}

\paragraph{Past-aware.} If we relax the complete independence assumption and allow the previous discourse structure to influence the generation of the current sentence, the Equation~\ref{equ:cond-prob} will be simplified as below. The discourse context in the instruction template includes only previous discourse sequence but not the future.
\begin{align*}
   P(\boldsymbol{x}|\boldsymbol{c}, \mathcal{I}) \approx & \prod_{i=1}^{|\boldsymbol{x}|} \prod_{j=1}^{|x_i|} p(x_{i,j} | x_{i,<j}, \boldsymbol{x}_{<i}, c_{\leq i}, \mathcal{I})
\end{align*}

\paragraph{Full-structure.} If we make no assumption and provide the full discourse structure, the Equation~\ref{equ:cond-prob} should be expressed as below.
Articles generated under different discourse information exposure are compared to determine the optimal instruction template. Experimental results are presented in Section~\ref{section:results}.
\begin{align*}
    P(\boldsymbol{x}|\boldsymbol{c}, \mathcal{I}) = & \prod_{i=1}^{|\boldsymbol{x}|} \prod_{j=1}^{|x_i|} p(x_{i,j} | x_{i,<j}, \boldsymbol{x}_{<i}, \boldsymbol{c}, \mathcal{I})
\end{align*}

In the input information section, we specify the input prompt of the overall generation task and the discourse role of the current generation unit.
For example, in Figure~\ref{fig:overview}, the instruction shown is for generating news report, where the input prompt is the news headline.
In the case of the recipe domain, dish title and ingredient list serve as the input, populating the corresponding template.

In the final component, we incorporate textual context to guide the generator in continuing writing the current text segment.
During SFT, preceding segments of the article up to the current target one are aggregated to form the previous text, while for inference, all previously generated texts are used.

\begin{table}[]
\centering
    \renewcommand{\arraystretch}{1}
    \scalebox{0.90}{
    \begin{tabular}{l} \hline
    {\fontsize{11}{14}\selectfont \textbf{Instruction template}}\\ \hline
        \begin{tabular}[c]{@{}l@{}}
            {\fontsize{11}{12}\selectfont The previous discourse structure is :}\\ 
            {\fontsize{11}{12}\selectfont \colorbox{mint}{\raisebox{0pt}[0.6em][-0.1em]{\textless{}Main Event\textgreater\ \textless{}Main Event\textgreater}}}\\ 
            {\fontsize{11}{12}\selectfont The future discourse structure is:}\\ 
            {\fontsize{11}{12}\selectfont \colorbox{mint}{\raisebox{0pt}[0.6em][-0.1em]{\textless{}Journalist Evaluation\textgreater\ \textless{}Anecdotal Event\textgreater}\ {[}...{]}} }\\ 
            {\fontsize{11}{12}\selectfont Please continue writing a \colorbox{red}{\raisebox{0pt}[0.6em][-0.1em]{\textless{}Consequence\textgreater}} section} \\
            for the news article about \colorbox{yellow}{\raisebox{0pt}[0.6em][-0.1em]{"Storm Friederike batters}} \\ 
            {\fontsize{11}{12}\selectfont \colorbox{yellow}{\raisebox{0pt}[0.6em][-0.1em]{Germany with hurricane-strength winds"}}}\\ 
            {\fontsize{11}{12}\selectfont News report:}\\ 
            {\fontsize{11}{12}\selectfont \colorbox{green}{\raisebox{0pt}[0.6em][-0.1em]{The United Kingdom envisions a futuristic landscape}}}\\
            \colorbox{green}{\raisebox{0pt}[0.6em][-0.1em]{in the year 2100, featuring state-of-the-art spaceports}}\\ 
            {\fontsize{11}{12}\selectfont \colorbox{green}{\raisebox{0pt}[0.6em][-0.1em]{and innovative sky farms as part of its evolving}} }\\
            \colorbox{green}{\raisebox{0pt}[0.6em][-0.1em]{infrastructure.}} {[}...{]} \\ \hline
        \end{tabular}
    \end{tabular}
    }
    \caption{Instruction example. The \colorbox{mint}{discourse context}, \colorbox{red}{current discourse role} and \colorbox{yellow}{headline} are dynamically adjusted based on the context and position of the current sentence. The \colorbox{green}{textual context} is all previous text before the current target sentence.}
    \label{tab:instruction_template}
    \vspace{-3mm}
\end{table}

\subsection{Zero-shooting LLMs}
While we fine-tuned LMs with above-mentioned instructions as the backbone generators of our sequential framework.
However, we also explored the option of using zero-shot prompting LLMs with minor modifications to the instruction template.
Specifically, for the SFT paradigm, we fine-tuned Flan-T5-base~\citep{chung2022scaling} and GPT-2 base \citep{radford2019language}.
In the case of the zero-shot setup, we opted GPT-3.5-turbo and Flan-T5-xxl.
These models have exhibited strong performance in general tasks but are either expensive or not readily available for further training. 

To enhance the LLMs' comprehension of the discourse schema, we introduced a natural langauge definition of the target discourse role at the beginning of the instruction template. Further details on the discourse definition are listed in Section~\ref{appen:schema}.
In Section~\ref{section:results}, we present the results achieved using backbone generators under both fine-tuned and zero-shot paradigms.
The results demonstrate the effectiveness and applicability of our framework across different settings.

\subsection{Measuring the discourse structure}

\begin{figure*}[ht!]
\begin{center}
\includegraphics[width=0.9\linewidth]{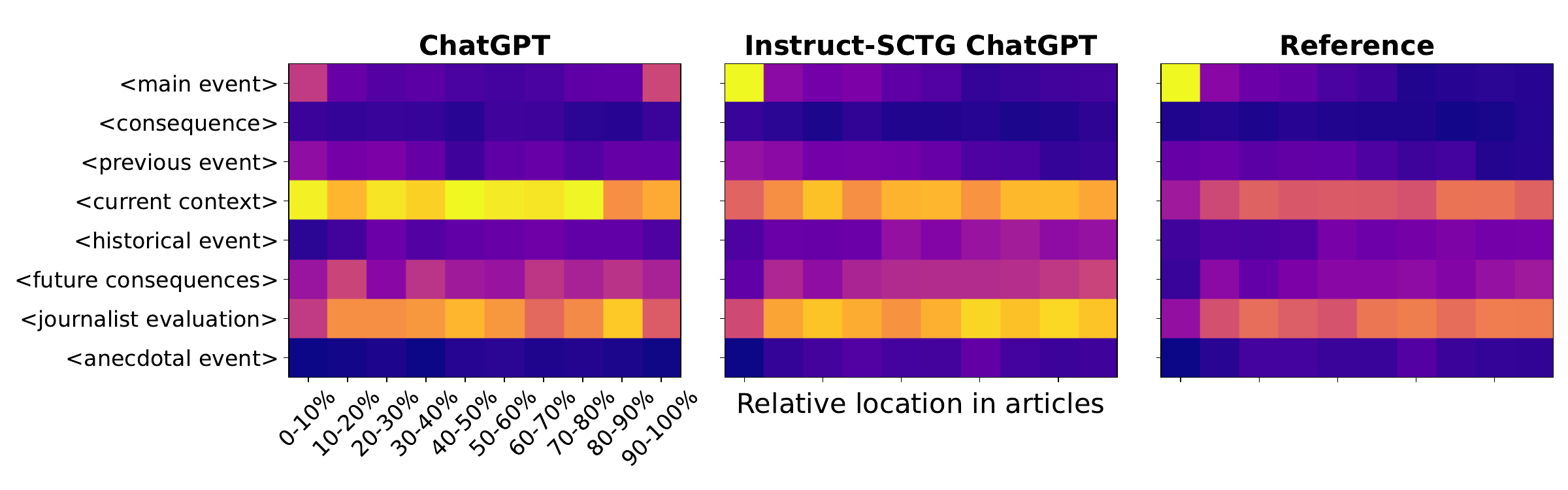}%
\end{center}
\caption{Comparison of discourse distributions at each relative position within news articles. The x-axis represents the relative position from the beginning to the end (0-9), while the y-axis represents different discourse roles based on the news schema of \citet{choubey-etal-2020-discourse}. Our framework (mid) demonstrates closer discourse distributions to the human-written articles (right), compared with the vanilla baseline (left).}
\label{fig:pos}
\vspace{-5mm}
\end{figure*}

Intuitively, for texts of a certain genre, they tend to follow similar discourse sequences while allowing for some degree of local flexibility. 
In other words, the distributions of discourse roles in similar areas of the articles are expected to be roughly similar. 
For instance, in news reports, it is common to have a sentence introducing the main event or consequence at the beginning to quickly capture readers' attention, but the exact position may vary.
In Figure~\ref{fig:pos}, we present the disparity between the discourse distributions of the articles generated by the zero-shot LLM and the reference texts written by humans is evident.

Therefore, to measure the positional difference between the discourse distributions in a fuzzy manner, we introduce the Positional Divergence $D_{pos}$ as an automatic metric.
Equation~\ref{equ:pos_div} demonstrates the calculation of the Positional Divergence. 
\begin{equation}
    D_{pos} = \frac{1}{N} \sum_{n=1}^{N} D_{KL}(p^n(r)||q^n(r))      \label{equ:pos_div}
\end{equation}

Here, $p^n(r)$ represents the distribution of discourse role $r$ for the reference data in the $n$-th position bin and $q^n(r)$ represents the distribution for the generated articles.
To compute this metric, we firstly segment the reference and generated articles from the evaluation set into $N$ bins based on their relative positions in the articles. 
Then, for each bin $n$, we calculate the KL divergence $D_{KL}(p^n(r)||q^n(r))$ between the discourse distributions with add-one smoothing to avoid zero probabilities.

Because the divergence is calculated based on their relative positions in the articles, it mitigates the impact of variations in segmentation styles or the total number of sentences, which cannot be solved by simply calculating the exact match rate. We further elaborate the difference and show that our positional divergence has high correlation with human evaluations in Section~\ref{appen:metric}.
We note that, for this metric, a discourse role classifier is required to label the generated articles.

\section{Dataset and Schema} \label{sec:schema}

In this work, we demonstrate the application of  our framework in two representative domains: News and Recipe.
News generation is considered an open-ended task, where there is no fixed predefined answer, allowing more room for creative variations.
Whereas Recipe generation is regarded as a closed-ended task, where there exists a correct reference recipe for a given input title.
To create the training data, we segment articles into sentences and label them with the assistance of discourse role classifiers. 

For news domain, we adopt an existing theory of functional discourse schema proposed by \citet{van1988news, van2013news}, which defines a discourse schema based on eight types of relations between each sentence and the main event.
A recent News Discourse dataset \citep{choubey-etal-2020-discourse} is manually annotated following the functional discourse schema, which contains 802 documents spanning over four domains and three media sources.
We utilize the training set of this dataset to train our discourse role classifier and the test set for evaluating the performance of our framework.
In addition, we label the Kaggle All-The-News dataset using our trained discourse role classifier, creating silver-labelled data.
Our backbone generators are fine-tuned on the All-The-News training set and evaluated on the News Discourse test set and All-The-News validation set.

For the domain of Recipe, we adopt the discourse schema proposed by \citet{liu-etal-2022-plug} which includes seven discourse roles based on cooking actions specifically designed for recipes.
We re-implement their discourse role classifier trained on a subset of the Recipe1M+ validation set \citep{marin2019learning}, where the discourse annotations are generated using a rule-based system.
We apply this classifier to the remaining Recipe1M+ dataset to generate the silver discourse labels. 
The fine-tuning of backbone generators for the Recipe domain is performed on the Recipe1M+ training set, and the evaluation is conducted on the Recipe1M+ test set.
Before using these datasets, we apply pre-processing and filtering based on specified conditions, as elaborated in Appendix~\ref{appen:preprocess}. 
For evaluation, we randomly sample 200 examples from each evaluation set to assess the performance of our framework and the baseline models, and the results are reported in Table~\ref{Tab:results_news} and \ref{Tab:results_recipes}.

\begin{table*}[ht]
    \centering  
    \renewcommand{\arraystretch}{1}
    \setlength{\tabcolsep}{3pt}
    \scalebox{0.90}{
        \begin{tabular}{llccc@{\hspace{8pt}}ccc@{\hspace{12pt}}cccc@{\hspace{8pt}}ccc}
        \hlinewd{0.8pt}
        \addlinespace[0.5ex]
        \multicolumn{2}{c}{\multirow{3}{*}{\textbf{Model}}} & \multicolumn{6}{c}{\textbf{Kaggle All-the-news}} & & \multicolumn{6}{c}{\textbf{News Discourse}}\\[1mm]
        \cline{3-8} \cline{10-15} 
        \multicolumn{2}{c}{} & \multicolumn{3}{c}{Fluency} & \multicolumn{3}{c}{Structure} &  & \multicolumn{3}{c}{Fluency} & \multicolumn{3}{c}{Structure} \\[0.5mm]
        \multicolumn{2}{c}{} & \textbf{PPL.}$\downarrow$    & \textbf{R-L}$\uparrow$    &\textbf{C-F}$\uparrow$    &\textbf{Acc.}$\uparrow$  &\textbf{Pos.}$\downarrow$    &\textbf{C-S}$\uparrow$    &  
        &\textbf{PPL.}$\downarrow$   &\textbf{R-L}$\uparrow$   &\textbf{C-F}$\uparrow$   &\textbf{Acc.}$\uparrow$  &\textbf{Pos.}$\downarrow$    &\textbf{C-S}$\uparrow$ \\[1mm]
        \hline
        \addlinespace[1ex]
        F-T
            & GPT2\textsubscript{Base} 
                & 87.8 & 21.1 & 2.7 & 25.3    &    0.31     &     3.0     & 
                & 91.2 & 19.7 & 3.1 & 20.2    &    0.36     &     2.2 \\
            & FT5\textsubscript{Base} 
                & 100.4  & 21.4 & 2.4 & 24.9 & 0.37 & 2.7 &
                & 108.1  & 20.4 & 2.9 & 21.3 & 0.32 & 2.4 \\[1mm]
        CTG  
            & GPT2\textsubscript{Base}  
                & 80.3  & 22.8 & 3.0 & 46.8 &   0.16 & 3.2 &
                & 86.3   & 20.4 & 3.2 & 44.9 &   0.18  & 2.9 \\[1mm]
        Z-S
            & GPT3.5 
                &5.9  & 22.7  &  4.7  &  35.2  &  0.19  & 4.0  & 
                & \underline{\textbf{4.9}}    &  21.6 & 4.4  & 32.5  &  0.21   &  4.4        \\
            & FT5\textsubscript{XXL}  
                & 10.0  & 22.4  &  4.2  &  30.6   & 0.20    &   3.9       &  
                & 11.8  & 21.0  & 4.3   &  36.1  &  0.22 &4.2 \\[1mm]
        \hline
        \addlinespace[0.2ex]
        \small{\textbf{I-SCTG}} \\[-1mm]
        {\hspace{4pt} F-T}
            & FT5\textsubscript{Base}-L   
                & 78.6  & 22.2   &   3.1   &  60.0    &  0.10     &   3.6   &
                & 74.5   &  21.8    &  3.3        &  63.2        &  0.13     &  3.6 \\
            & FT5\textsubscript{Base}-P 
                & 63.5 &  \underline{\textbf{23.5}}   &  3.2  & \underline{\textbf{63.5}} & \underline{\textbf{ 0.08}}    & 3.7     & 
                & 65.1   & 21.1     &  3.5   & \underline{\textbf{67.6}} & 0.10 & 3.2 \\
            & FT5\textsubscript{Base}-F 
                & 65.1 &  22.4   &  3.5   &   61.4  &  0.11    &  3.5     &  
                & 67.9  & 20.9    &   3.3   &  65.7 & \underline{\textbf{0.09}} & 3.5 \\[1mm]
        {\hspace{4pt} Z-S}
            & GPT3.5-P    
                & \underline{\textbf{5.7}}  & 22.4       & \underline{\textbf{4.8}} & 48.5 &  0.16 &  \underline{\textbf{4.8}} &  
                &  6.7 & 22.5       &  \underline{\textbf{4.5}}     & 40.0  &   0.17   & \underline{\textbf{4.7}}    \\
            & FT5\textsubscript{XXL}-P    
                & 9.1        &  22.6      &   4.4    & 42.1 & 0.18 &   4.3   &  
                & 10.8 & \underline{\textbf{23.1}}  &   4.2  &  42.5   &    0.19   &   4.4 \\ 
        \hlinewd{0.8pt}
        \end{tabular}
    }   
    \caption{Results of automatic evaluations conducted on the News domain. The top half shows the outcomes for three types of baseline methods, while the bottom half for various model settings within our Instruct-SCTG (I-SCTG) framework. In the table, "L" denotes the setting for local-discourse, "P" for past-aware, and "F" for full-structure. 
    Our framework shows better ability in controlling the discourse structure of the generated text. For fine-tuned backbone generators, our framework also achieves better surface fluency.}
    \label{Tab:results_news}
\vspace{-3mm}
\end{table*}

\section{Experiments}
\subsection{Implementation Details}
In the news domain, the Flan-T5-base backbone generator is trained on the Kaggle All-The-News pre-processed training set for 200k steps, using a batch size of 4.
For recipe domain, training is conducted on the processed Recipe1M+ training set for 100k steps with a batch size of 8.
Both generators are optimized using the Adam optimizer~\citep{kingma2014adam} with a learning rate of $3e-5$ and an L2 decay rate of 0.05.
As for the zero-shot backbone generators, we employ the GPT-3.5-turbo (ChatGPT) and Flan-T5-xxl. 
During inference, a temperature value of 0.7 is set for news generation and 0.2 for recipes.
For news generation, we utilize the top-p sampling method with a value of $p=0.8$, while for recipe generation, we employ beam search decoding with a beam size of 5.

Regarding the Flan-T5, it has limits on the maximum sequence length for both input and output. 
Therefore, we truncate the input textual context from the beginning to ensure the instruction prompt does not exceed 1024 tokens. 
The maximum output length is set at 256 tokens.


For the discourse role classifiers, we fine-tune a DistilBERT model~\citep{sanh2019distilbert} using the News Discourse training set for the news domain and the recipe1M+ training set for recipes.
Both classifiers are trained for 10k steps with a batch sizes of 32.
The remaining hyper-parameters are the same with the settings of the backbone generators.
The DistilBERT model, being relatively lightweight, demonstrates promising performance as discussed in Appendix~\ref{sec:result_cls}.

\subsection{Experimental setup}
\noindent\textbf{Metrics}\quad
We assess our framework from two main perspectives: Surface fluency and adherence to the discourse structure.
To measure the surface fluency, we utilize established metrics such as BLEU (\textbf{B})~\citep{papineni-etal-2002-bleu}, ROUGE-L \textbf{R-L}~\citep{lin-2004-rouge} and perplexity \textbf{(PPL.}) by another language model OPT-2.7B~\citep{zhang2022opt}. 
As for discourse structure control, we measure the exact match accuracy (\textbf{Acc.}), which is the average percentage of matched discourse sequences between the generated text and the reference.
Additionally, we use the previously described positional discourse divergence (\textbf{Pos.}) with the number of bins $N=10$.

Traditional automatic metrics often struggle to capture inter-sentence coherence, especially in open-ended generation tasks.
Following a recent work~\citep{kocmi2023large}, we employ ChatGPT to perform evaluation on both textual fluency (\textbf{C-F}) and structural coherence (\textbf{C-S}) with a scale from $1$ to $5$.
Furthermore, we also perform human evaluations on these aspects by hiring three native English speakers.
They evaluate a randomly selected subset of 100 examples for each evaluation dataset, producing ratings on a scale of $1$ to $5$.
Detailed information about the evaluation prompts for ChatGPT and the setup for human evaluation can be found in Appendix~\ref{appen:chat} and \ref{appen:human}.


\noindent\textbf{Baselines}\quad
We evaluate our framework against three types of baselines:
1) Vanilla Fine-tuned LMs (\textbf{F-T}): We fine-tune a GPT-2-base and a Flan-T5-base using only input headlines and reference text pairs from the All-The-News and Recipe1M+ training sets, without incorporating discourse information. We employ the top-k sampling decoding method with a value of $k=5$.
2) Controlled Text Generation methods (\textbf{CTG}): We compare against the approach proposed by \citet{liu-etal-2022-plug}, which utilizes a discourse classifier to guide the decoding of a fine-tuned GPT-2-base backbone decoder.
3) Zero-shot large language models (\textbf{Z-S}): We experiment with the GPT-3.5-turbo and Flan-T5-xxl models, prompting them only with the input but no discourse information. The remaining hyper-parameters remain consistent with our framework.

\begin{table}[]
\centering  
\renewcommand{\arraystretch}{1.2}
   \scalebox{0.9}{
        \begin{tabular}{lcc}
        \hlinewd{0.8pt}
        \textbf{Model}  & \textbf{Fluency} & \textbf{Coherence} \\ \hline
        FT5\textsubscript{Base}-FT & 2.4              & 2.5                \\
        GPT3.5-ZS       & \underline{\textbf{4.5}}     & 4.0                \\
        \hline
        FT5\textsubscript{Base}-P  & 2.5              & 3.5                \\
        GPT3.5-P          & 4.3              & \underline{\textbf{4.2}}       \\   
        \hlinewd{0.8pt}
        \end{tabular}
    }
    \caption{Results of human evaluations on the News Discourse test set comparing baselines with our Instruct-SCTG framework. Our framework demonstrates improved structural coherence while maintaining a comparable level of surface fluency.}
    \label{tab:human}
\vspace{-5mm}
\end{table}

\subsection{Results}\label{section:results}
\subsubsection{News articles}
Experimental results were obtained using a randomly selected subset of 200 samples for each dataset. 
Table~\ref{Tab:results_news} displays the averaged experimental results over 5 runs with different random seeds for news generation using various methods. 

The results demonstrate that our framework outperforms all baseline models on surface fluency and structural coherence metrics when using fine-tuned backbone generators. 
Among the different contextual discourse information settings, past-aware exhibits better performance. 
This could be attributed to the fact that subsequent discourse structures might not provide informative enough guidance and could distract the attentions from the more important current discourse roles. 
When employing zero-shot generators, our framework only utilizes past-aware discourse structure setup to minimize the computational cost. 
Although vanilla zero-shot LLMs achieve satisfactory surface fluency, our framework can still further enhance the structural coherence of the generated text.


In terms of human evaluations, our framework is compared to two representative baseline models, and the results are presented in Table~\ref{tab:human}. 
The human evaluations align with the findings from automatic metrics, confirming that our Instruct-SCTG framework can effectively control the generation process to adhere to the provided discourse structure, resulting in improved structural coherence, while maintaining comparable surface fluency.

\begin{table}[t!]
    \centering  
    \renewcommand{\arraystretch}{1}
    \setlength{\tabcolsep}{3pt}
    \scalebox{0.9}{
        \begin{tabular}{llccc@{\hspace{6pt}}cc}
        \hlinewd{0.8pt}
        \addlinespace[0.5ex]
        \multicolumn{2}{c}{\multirow{3}{*}{\textbf{Model}}} & \multicolumn{5}{c}{\textbf{Recipe1M+}} \\[1mm]
        \cline{3-7} 
        \multicolumn{2}{c}{} & \multicolumn{3}{c}{Fluency} & \multicolumn{2}{c}{Structure} \\[0.5mm]
        \multicolumn{2}{c}{} & \textbf{B}$\uparrow$ & \textbf{PPL.}$\downarrow$    & \textbf{R-L}$\uparrow$    & \textbf{Acc.}$\uparrow$  & \textbf{Pos.}$\downarrow$    \\ [1mm]
        \hline
        \addlinespace[1ex]
        F-T
            & GPT2\textsubscript{Base} &  13.1   &     28.1    &   38.0  &  29.3 & 0.36 \\
            & FT5\textsubscript{Base} &  12.7   & 27.4 & 37.3 & 27.9 & 0.41 \\[1mm]
        CTG  
            & GPT2\textsubscript{Base}  &  15.8   & 26.8 & 39.1 & 50.8 & 0.14 \\[1mm]
        Z-S
            & GPT3.5  & \underline{\textbf{19.2}} & 7.7  & \underline{\textbf{44.5}}  &  35.2        & 0.25 \\
            & FT5\textsubscript{XXL}  &  17.7 &  9.5 & 43.2&  32.3  & 0.27  \\[1mm]
        \hline
        \addlinespace[0.2ex]
        \small{\textbf{I-SCTG}} \\[-1mm]
        {\hspace{4pt} F-T} 
            & FT5\textsubscript{Base}-L   &  16.5  &  19.8   &   40.3   &  66.0    &  0.10  \\
            & FT5\textsubscript{Base}-P & 16.3  &  24.0  &   40.5   &   \underline{\textbf{68.3}} & \underline{\textbf{0.08}}       \\
            & FT5\textsubscript{Base}-F &   15.8  &  23.2 &  39.8  & 67.5 &  \underline{\textbf{0.08}}    \\[1mm]
        {\hspace{4pt} Z-S} 
            & GPT3.5-P    & 19.0  &  \underline{\textbf{6.1}}  &  44.2  &  47.6     &   0.15   \\
            & FT5\textsubscript{XXL}-P    &   18.1 &   8.2   &  43.5   &  49.2   &   0.15   \\ 
        \hlinewd{0.8pt}
        \end{tabular}
    }   
    \caption{Automatic evaluation results for the Recipe domain. Our framework exhibits excellent performance in controlling discourse structure. Improvements in textual fluency are observed when applied our framework to the fine-tuned generators.}
    \label{Tab:results_recipes}
\vspace{-3mm}
\end{table}

\subsubsection{Recipes}
Having the same experiment setup as the news domain, we present the results in Table~\ref{Tab:results_recipes}.
We observe similar trend with the results on news datasets: Our framework improves the structure coherence for both types of  generators, while only fine-tuned generators exhibit better surface fluency.
This can be attributed to fact that the recipes generated by latest large-scale LMs already achieve satisfactory fluency, leaving limited room for further improvement.
By applying our framework, the order of actions can be adjusted to better align with the input discourse sequence, while the fluency level remains comparable due to the strong generation capabilities of LLMs.
On the other hand, for the fine-tuned generators, incorporating more natural discourse structures can effectively enhance fluency.



\section{Conclusion}

In this work, we address the task of controlling the discourse structure during the generation process.
We propose a sequential framework, the Instruct-SCTG, which decomposes article generation into sentence-level tasks. 
Our framework effectively leverages supervised fine-tuned LMs or zero-shot LLMs as backbone generators to produce structurally more coherent text.
We also propose the automatic metric, positional discourse divergence, measuring the discrepancy in discourse distributions across relative positions within the articles.
Extensive evaluations demonstrate that our framework can effectively leverage instruction-following LMs to align the discourse structures and achieve SOTA performance on SCTG tasks in both News and Recipe domains.

\section*{Limitations}

\paragraph{Hallucination of news content} 
In our experimental setup, our primary focus is on controlling the discourse structure of the generated text, rather than the content itself. Consequently, there is a potential for hallucination or the generation of inaccurate information. We acknowledge that in the domain of news reports, the presence of unfactual content can pose problems for readers, as it may compromise the credibility and reliability of the generated articles.

\paragraph{Length limitations}
News articles are typically lengthy, but current LLMs often have constraints on maximum input or output token length. We acknowledge that the truncation method employed in our study may not be optimal, and alternative approaches for encoding/decoding extra-long articles could be explored to capture more contextual information.

\paragraph{Granularity of discourse annotations} 
When applying our framework on the zero-shot backbone generators, we observe instances of local repetition where consecutive sentences conveyed similar meanings. 
This may be attributed to the LLMs' differing understanding of the granularity of discourse structure compared to the reference annotations.
LLM-generated articles tend to have fewer sentences, resulting in shorter discourse sequences. We recognize that this issue could potentially be improved by employing more suitable granularity when annotating the discourse labels.

\paragraph{Data leakage in LLMs}
Modern LLMs use enormous corpora during pre-training stage, some of which may not be publicly disclosed. 
News data, in particular, has a tendency to be easily accessible, because for an event there might be multiple source of reporting, which makes them easily scraped for the pre-training.
As a result, experiments conducted on news datasets may not be as indicative as before due to potential data leakage concerns.


\bibliography{anthology,custom}
\bibliographystyle{acl_natbib}

\appendix

\section{Appendix}
\label{sec:appendix}
\subsection{Discourse Classifier Results}\label{sec:result_cls}
For the News domain, the discourse role classifier is trained on the News Discourse training set and evaluated on the validation set using human-annotated gold labels. The classifier achieves an accuracy of $67\%$.

In the Recipe domain, the discourse role classifier is trained on the Recipe1M+ training set and evaluated on the validation set using silver annotations generated by the rule-based system proposed by \citet{liu-etal-2022-plug}. The classifier achieves an accuracy of $92\%$.

\subsection{ChatGPT Evaluation Templates} \label{appen:chat}
We use the following instruction to prompt the ChatGPT to rate the textual fluency \textbf{C-F} and structural coherence \textbf{C-S} of the generated texts.

\enquote{You are a helpful virtual journalist. Please rate the textual fluency of the following news report with a score from 1 to 5. Only return the value:}

\enquote{You are a helpful virtual journalist. Please rate the structural coherence and the discourse structure quality of the following new report with a score from 1 to 5. Only return the value:}

\subsection{Human Evaluation Guidance Questions} \label{appen:human}
Please rate the following article from two aspects: 1) Textual fluency and 2) structural coherence with score 1 to 5. When evaluating the article, please consider the following guidance.
\begin{itemize}[noitemsep,topsep=1pt]
    \itemsep 0em
    \item \textbf{Introduction and lead}: Does the article have a clear and engaging introduction that effectively presents the main topic and captures the reader's attention?
    \item \textbf{Structure organizatio}: Do the sections and paragraphs follow a clear structure that contributes to the overall understanding of the topic? Are the paragraphs well-structured, with clear topic sentences and appropriate supporting details? Do the paragraphs transition smoothly, maintaining a consistent flow of ideas?
    \item \textbf{Clarity and precision}: Is the language clear, concise, and precise? Are the ideas expressed in a way that is easy to understand for the target audience?
    \item \textbf{Use of evidence and sources}: Are relevant sources and evidence used to support the article's claims and arguments?
\end{itemize}

\subsection{Discourse Schema}
\label{appen:schema}
The definition of the discourse schema we used for news articles:
\begin{itemize}[noitemsep,topsep=1pt]
    \itemsep 0em
    \item \textbf{Main Event}: The major subject of the news article.
    \item \textbf{Consequence}: An event or phenomenon that is caused by the main event.
    \item \textbf{Previous Event}: A specific event that occurred shortly before the main event.
    \item \textbf{Current Context}: The general context or world state immediately preceding the main event.
    \item \textbf{Historical Event}: An event occurring much earlier than the main event.
    \item \textbf{Future Consequences}: An analytical insight into future consequences or projections.
    \item \textbf{Journalist Evaluation}: A summary, opinion or comment made by the journalist.
    \item \textbf{Anecdotal Event}: An event that is uncertain and cannot be verified. The primary purpose is to provide more emotional resonance to the main event.
\end{itemize}

\noindent The definition of the discourse schema we used for recipes:
\begin{itemize}[noitemsep,topsep=1pt]
    \itemsep 0em
    \item \textbf{Pre-processing} means the preparations of ingredients or cooker.
    \item \textbf{Mixing} includes actions of combining one or more ingredients together.
    \item \textbf{Transferring} is for the actions of moving or transferring food or intermediate food to a specific place.
    \item \textbf{Cooking} represents the actual cooking actions, which could vary drastically across different recipes.
    \item \textbf{Post-processing} usually refers to the following up actions after the `cooking' stage, such as `cooling down', `garnish'.
    \item \textbf{Final} refers to the last few actions before serving the food or the serving action itself.
    \item \textbf{General} includes the rest of actions which cannot be classified into the above categories.
\end{itemize}

\subsection{Further details on Positional Divergence }\label{appen:metric}
\textbf{Metric Necessity.} We clarify two main practical benefits of our proposed metric:
\begin{itemize}[noitemsep,topsep=1pt]
\itemsep 0em
\item     For open-ended generation tasks, \textbf{it is common for the generated text to have different length (different total number of sentences) or different paragraph layout (different number of sentences for each paragraph) as compared to reference text}. However, these variations do not necessarily mean a substantial deviation in discourse structure. To address this, our proposed positional divergence focuses only on comparing discourse role distributions based on the corresponding relative positions. The continual labels merging strategy couldn’t provide a correct paragraph segmentation due to the aforementioned discontinuity.
\item The discourse labels for \textbf{the existing dataset don’t usually have multi-sentence continuity, either because the labels are noisy or the flexible nature of the text from open-ended domains}. For instance, below we show the discourse labels for the sentences in the first paragraph of the Number 18 datapoint of the News Discourse dataset test set: [‘main’, ‘previous\_event’, ‘main’, ‘journalist\_evaluation’, ‘main’, ‘main’, ‘main’, ‘main’, ‘main’, ‘consequence’]. While the main role of the paragraph is to describe the <main event>, sentences within it might be assigned different role labels such as <evaluation> or <consequence>. In such cases, a simplistic strategy like merging continual sentences cannot effectively handle the evaluation unless guided by a sophisticated merging policy.

\end{itemize}

\textbf{Metric Effectiveness.} We conducted supplementary evaluations to further justify the effectiveness of our metric. We compare the correlation of our metric and the exact match rate to the human evaluation results. In Table~\ref{tab:corr}, we show correlations on the same 100 examples from the News Discourse dataset as shown in Table~\ref{tab:human}. The results show that \textbf{our positional divergence has generally higher correlations than the exact match}. 
\begin{table}[]
\centering
\begin{tabular}{l|c|c|}
\cline{2-3}
                                   & $\rho$(\textbf{Acc}., H.C.) & $\rho$(\textbf{Pos.}, H.C.) \\ \hline
\multicolumn{1}{|l|}{FT5\textsubscript{base}-FT} & 0.19                               & 0.32        \\ \hline
\multicolumn{1}{|l|}{FT5\textsubscript{base}-P}  & 0.28                               & 0.36        \\ \hline
\multicolumn{1}{|l|}{GPT3.5-ZS}    & 0.26                               & 0.33        \\ \hline
\multicolumn{1}{|l|}{GPT3.5-P}     & 0.24                               & 0.36        \\ \hline
\end{tabular}
\caption{The correlations between Human Coherence (H.C) and Exact Match (Acc.) and between H.C. and Positional Divergence. Our proposed metric has shown better correlation with human evaluation. }
\label{tab:corr}
\end{table}

\subsection{Data Preprocessing}
\label{appen:preprocess}
For Kaggle All-The-News, we filtered the dataset based on the following conditions:
\begin{itemize}[noitemsep,topsep=1pt]
    \itemsep 0em
    \item Containing special characters: @, [, +.
    \item Having total number of words over 800 or below 100.
    \item Containing random comments.
    \item Containing more than two reports.
\end{itemize}
Then we pre-process the data by
\begin{itemize}[noitemsep,topsep=1pt]
    \itemsep 0em
    \item Removing extra space.
    \item Removing reporting source.
    \item Removing journalist names.
    \item Removing emoji.
\end{itemize}

\noindent For Recipe1M+, we filter it based on the following codintions:
\begin{itemize}[noitemsep,topsep=1pt]
    \itemsep 0em
    \item Containing irrelevant information, such as advertisements, reviews and comments.
    \item Having total number of words over 300 or below 50.
    \item Duplicate recipes.
\end{itemize}

\end{document}